\documentclass[11pt,conference]{IEEEtran}
\usepackage{cite}
\usepackage{amsmath,amssymb,amsfonts}
\usepackage{algorithm}
\usepackage{algorithmic}
\usepackage{graphicx}
\usepackage{textcomp}
\usepackage{xcolor}
\usepackage{float}
\usepackage{booktabs} % For professional table formatting
\IEEEoverridecommandlockouts

\title{A Blockchain-Monitored Agentic AI Architecture for Trusted Perception--Reasoning--Action Pipelines%
\thanks{This paper was presented at the IEEE International Conference on Computing and Applications (ICCA 2025), Bahrain.}}

\begin{document}

%	\title{A Blockchain-Monitored Agentic AI Architecture for Trusted Perception–Reasoning–Action Pipelines}

\author{
	\IEEEauthorblockN{Salman Jan\IEEEauthorrefmark{1}, Hassan Ali Razzaqi\IEEEauthorrefmark{1}, Ali Akarma\IEEEauthorrefmark{2}, Mohammad Riyaz Belgaum\IEEEauthorrefmark{1}}
	
	\IEEEauthorblockA{\IEEEauthorrefmark{1}Faculty of Computer Studies, Arab Open University-Bahrain }
	
	\IEEEauthorblockA{\IEEEauthorrefmark{2}Faculty of Computer and Information System, Islamic University of Madinah, Saudi Arabia}
	
	\IEEEauthorblockA{*Corresponding author: salman.jan@aou.org.bh}
}
	\maketitle
	
	\begin{abstract}
		The application of agentic AI systems in autonomous decision-making is growing in the areas of healthcare, smart cities, digital forensics, and supply chain management. Even though these systems are flexible and offer real-time reasoning, they also raise concerns of trust and oversight, and integrity of the information and activities upon which they are founded. The paper suggests a single architecture model comprising of LangChain-based multi-agent system with a permissioned blockchain to guarantee constant monitoring, policy enforcement, and immutable auditability of agentic action. The framework relates the perception conceptualization-action cycle to a blockchain layer of governance that verifies the inputs, evaluates recommended actions, and documents the outcomes of the execution. A Hyperledger Fabric-based system, action executors MCP-integrated, and LangChain agent are introduced and experiments of smart inventory management, traffic-signal control, and healthcare monitoring are done. The results suggest that blockchain-security verification is efficient in preventing unauthorized practices, offers traceability throughout the whole decision-making process, and maintains operational latency within reasonable ranges. The suggested framework provides a universal system of implementing high-impact agentic AI applications that are autonomous yet responsible.
	\end{abstract}
	
	\begin{IEEEkeywords}
		Blockchain, Agentic AI, Security, Privacy, Healthcare, IoT, Digital Forensics.
	\end{IEEEkeywords}

\section{Introduction}

Combining blockchain with agentic artificial intelligence (AI) is becoming more important in the development of secure, reliable and autonomous systems. While agentic AI offers planning, constraint assessment, and tool execution, creating auditable perception-reasoning-action loops, blockchain offers immutability, consensus, provenance, and smart contracts.

The blockchain is commonly used in inter-organizational and integrity-sensitive settings and facilitates decentralized data sharing, biometric access, permissioned records, and secure cloud-IoT settings \cite{abutaleb2023_patientcentric_access,butt2022_secure_record_sharing,mastoi2025_healthcare_cloud,syed2025_cloud_aura}. It also facilitates chain-of-custody, evidence tracing and privacy preserving forensic analytics \cite{alqahtany2024_forensictransmonitor,alqahtany2024_mobile_vpn_forensics,alqahtany2025_theft_blockchain,jan2021_smartOS_blockchain_generative}. Other uses are water management, vehicle lifecycle, property registration, VAT systems, Takaful banking, and smart IoT trust zones  \cite{abdeen2019_takaful_blockchain,ali2019_smartIoT_trustzone,ali2020_blocktrackL,ali2020_property_registration,ali2020_vehicle_lifecycle_blockchain,mastoi2023_water_blockchain,syed2019_vat_blockchain}.

The agentic AI is used with autonomous planning and multiagent operations in the demand forecasting, the digital-twin modeling, the IoT-healthcare, and tailored support of disabilities and neurodivergence \cite{jan2025_simdte,syed2025_agentic_demand_forecasting, syed2025_iot_blockchain_vit}. Nevertheless, it has no verifiable guarantees, and blockchain does not have adaptive reasoning.

In this paper, an agentic AI architecture is proposed that is monitored through blockchain, and every perception-conceptualization-action cycle is stored and verified on a permissioned blockchain. LangChain supports multi-agent reasoning, smart contracts authenticate information, rules, and approved actions run across MCP systems in smart-city, healthcare, and enterprise systems. We report related literature, architecture, Hyperledger Fabric implementation, and analyses on critical areas.

\section{Background}

\textbf{A. Blockchain as an Enabler for Trusted Applications}

Blockchain offers integrity-critical, auditable, multi-stakeholder, and decentralized records. Authenticated blockchains facilitate safe cloud-IoT information interchange, patient-centric access controls, CRBAC/Z-notation systems \cite{abutaleb2023_patientcentric_access,mastoi2025_healthcare_cloud,shah2025_blockchain_powered_healthcare,syed2018_crbac_znotation,syed2025_intent_gan,jan2025disable}, and interoperability through HL7 and identity architecture frameworks \cite{syed2025_water_digitaltwins_transformers,syed2025cloudburst}.

Blockchain can be used in digital forensics to ensure verifiable chains of custody, as well as to extract trends \cite{alqahtany2024_forensictransmonitor,alqahtany2025_theft_blockchain,syed2025_deepfake_blockchain,syed2025_smartwater_digitaltwins}. On-chain metadata is used in mobile VPN forensics using deep learning \cite{alqahtany2024_mobile_vpn_forensics,syed2025_ncall_behaviour_attestation}, and generative models are used to conduct smart OS validation \cite{jan2021_smartOS_blockchain_generative}.

Other systems are IoT provenance \cite{ali2020_blocktrackL}, urban water monitoring \cite{mastoi2023_water_blockchain}, trust-zone IoT security \cite{ali2019_smartIoT_trustzone}, vehicle lifecycle tracking \cite{ali2020_vehicle_lifecycle_blockchain}, property registration \cite{ali2020_property_registration}, VAT audits \cite{syed2019_vat_blockchain}, Takaful banking \cite{abdeen2019_takaful_blockchain}, access-control extensions \cite{syed2019_comparative_blockchain,khan2019_accesscontrol_permissioned,khan2020_blocku}, and smart contracts for virtual tourism and IoT \cite{siddiqui2020_virtual_tourism_blockchain,syed2017_iot_smartcontracts}.

\textbf{B. Agentic AI and Its Application Landscape}

The agentic AI incorporates the large models, planning, memory, and tool use into the autonomous perception-conceptualization-action loops. They are used in demand forecasting \cite{syed2025_agentic_demand_forecasting} and time-critical decision-making simulations using digital-twin \cite{jan2025_simdte}. Prior work on IoT trust zones, provenance, and healthcare sharing \cite{abutaleb2023_patientcentric_access,ali2019_smartIoT_trustzone,ali2020_blocktrackL,mastoi2025_healthcare_cloud} informs sensor, conceptualization, and action agents interacting with APIs or smart contracts. Security is strengthened through behavioral attestation \cite{syed2012_sense_of_others}, permission analysis \cite{syed2020_android_permission_framework}, and intrusion-detection schemes \cite{awan2025_trustworthy_clustering_vehicular,rehman2025_talb_iot,shaikh2024_ddos_hybrid_vit}.

\textbf{C. Agentic AI in Applications}

Autonomous AI is done on medical contexts  (e.g., real-time monitoring \cite{syed2025aghealth}) and supply chains through predictive inventory systems \cite{syed2025inventory}.This is enhanced by blockchain which has verifiable, tamper-evident records of decisions, which enhance trust and auditability.

\section{Literature Review}

The application of blockchain in data security and privacy has been widely investigated.  For example, \cite{shah2025_blockchain_powered_healthcare} promotes the use of blockchain in secure healthcare, and \cite{ali2019_smartIoT_trustzone} concerns the issue of IoT security. Within the context of digital forensics, \cite{alqahtany2024_mobile_vpn_forensics} focuses on the use of blockchain in guaranteeing evidence integrity throughout the chain of custody.

In autonomous decisionmaking, agentic AI has been used in healthcare \cite{syed2025aghealth} and inventory optimization \cite{syed2025inventory}, where machine learning, deep learning, and data mining are used. Nevertheless, there are still problems with information security and credibility. These problems can be addressed by combining blockchain with agentic AI whereby AI inputs are auditable and immutable.

The recent research on the AI governance based on blockchain is primarily concerned with post-hoc verification. Our framework, in contrast, implements policy at the contract level as part of the agentic decision-cycle directly implementing validation rules into the perception-reasoning-action cycle. This closely integrates sovereignty with control, which explains the novelty of our strategy in blockchain-AI systems.

\section{Proposed Blockchain-Monitored Agentic AI Framework}
\label{sec:proposed-framework}

This section presents an agentic AI design that is authorized through a blockchain to make secure, auditable and responsible decisions. It is based on a three-agent stack (Perception, Conceptualization, Action) and a LangChain-based agent stack. Decisions with high impact are monitored on a blockchain layer of governance and implemented through the Model Context Protocol (MCP).

\subsection{Architecture Overview}

\begin{figure*}[t]
	\centering
	\includegraphics[width=\linewidth]{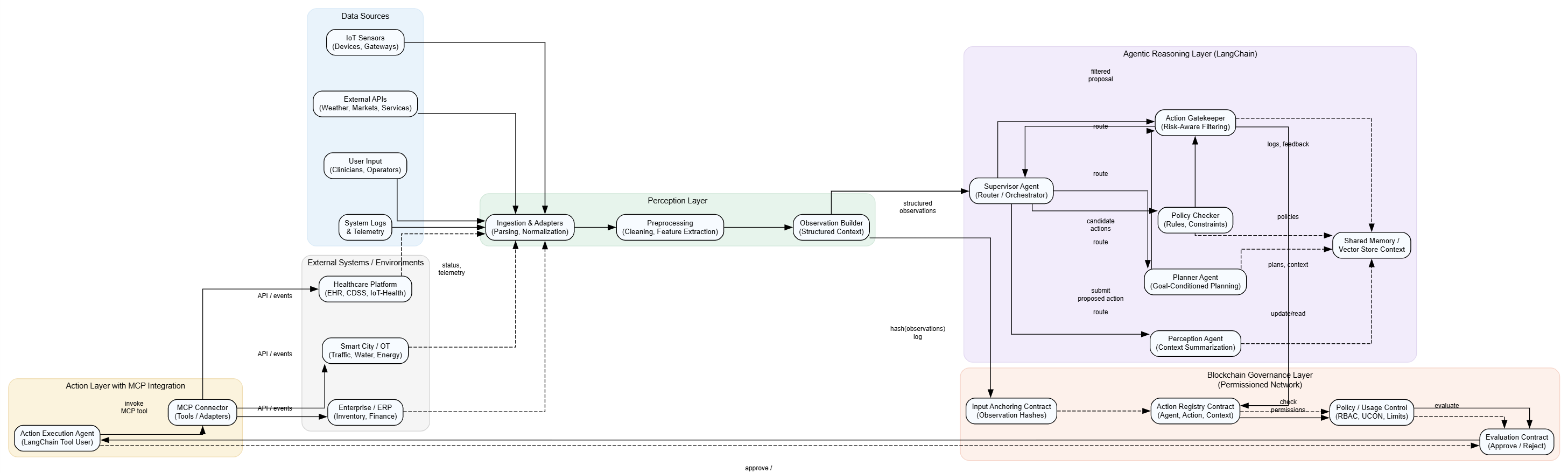}
	\caption{Proposed blockchain-governed agentic AI architecture integrating perception, LangChain-based reasoning, blockchain evaluation, and MCP-enabled action execution.}
	\label{fig:proposed-architecture}
\end{figure*}

The architecture (Figure~\ref{fig:proposed-architecture}) consists of four layers:

\begin{itemize}
	\item \emph{Perception Layer}: Gathers the raw data out of databases, IoT sensors, APIs and UIs to form structured observations.
	\item \emph{Conceptualization Layer}: LangChain agents can think, plan, and evaluate policies to come up with candidate actions.
	\item \emph{Blockchain Governance Layer}: Captures decision data, implements policies through smart contract execution, as well as provides audit trails, which cannot be tampered with.
	\item \emph{Action Layer with MCP}: Uses MCP-entitled tools that connect to ERP applications, smart city controllers, or healthcare infrastructures to carry out blockchain-approved actions.
\end{itemize}

The pipeline provides the entire observation-to-actuation chain can be proved by on-chain evidence by circumventing agentic AI with blockchain.

\subsection{Perception Layer: Secure Observation Ingestion}

The perception layer takes the input of various modalities (event logs, sensor outputs, user requests):

\[
\mathcal{O}_t = \{ o_t^1, o_t^2, \ldots, o_t^n \}.
\]
The processing of observations and low-level integrity checks are done by a Perception Agent. The anchoring of high-sensitivity inputs to the blockchain is through hashed \emph{input anchors} which connect original data to lower stream decisions.

\subsection{Conceptualization Layer: LangChain-Based Agentic Reasoning}

LangChain agents (planner, risk assessor, policy checker, explainer) take \(\mathcal{O}_t\) and:
\begin{enumerate}
	\item Generate candidate actions: \(\mathcal{A}_t = \{ a_t^1, \ldots, a_t^k \}\)
	\item Evaluate them under domain constraints, safety rules, and optimization objectives
	\item Select action \(a_t^\star\) for execution
\end{enumerate}

A Supervisor Agent sends a transaction proposal of the action, subset of observation hashes, agent identity, policy context to the blockchain to be verified, and executed.

\subsection{Blockchain Governance Layer: Monitoring and Evaluation}

A permissioned blockchain verifies, records, and enforces policy on the proposal of actions through smart contracts:

\begin{itemize}
	\item \emph{Action Registry Contract}: Records proposals and metadata.
	\item \emph{Policy and Usage Control Contract}: Imposes the safety rules, rate limits as well as RBAC permissions.
	\item \emph{Evaluation Contract}: Checks structure, policy, and regularity  of \(a_t^\star\)
\end{itemize}

For each proposal, the blockchain:
\begin{enumerate}
	\item Checks identity and data format of agent
	\item Guarantees compliance of roles and context
	\item Optionally triggers risk/compliance oracles
	\item Emits \emph{ActionApproved} or \emph{ActionRejected}
\end{enumerate}

Approved actions are transferred to MCP and all relationships of observation actions are stored permanently.

\subsection{Action Layer: MCP-Based Actuation}

The \emph{Action Execution Agent} subscribes to blockchain approvals and:
\begin{enumerate}
	\item Constructs an MCP request with specifications of target system and parameters.
	\item Translates it to API calls via MCP
	\item Gathers the external system response \(e_t\)
\end{enumerate}

Status codes, transaction IDs (key results) are hashed and anchored and fulfill the perception-action evidentiary loop.

\begin{figure}[t]
	\centering
	\includegraphics[width=\linewidth]{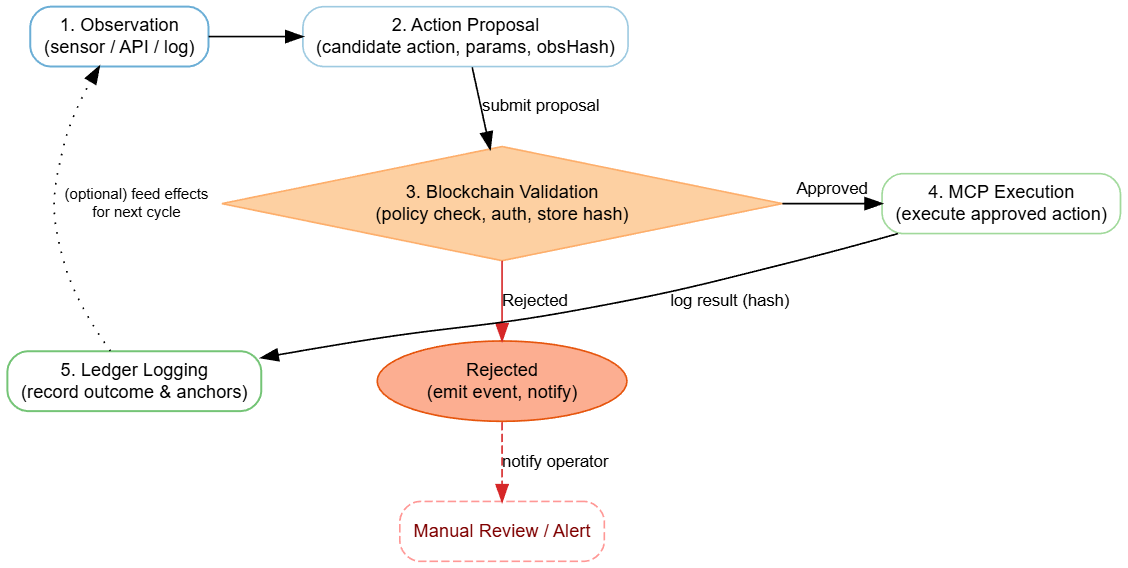}
	\caption{Layered agentic AI architecture governed by a permissioned blockchain, showing directional decision flow from observation to reasoning, blockchain validation, and MCP execution.}
	\label{fig:flow}
\end{figure}

\subsection{Blockchain Monitor Module as a LangChain Tool}

The \emph{Blockchain Monitor Module} allows the interaction of the first class with blockchain functions:

\begin{itemize}
	\item \texttt{log\_observation(id, hash, metadata)}
	\item \texttt{submit\_action(action\_spec)}
	\item \texttt{check\_status(action\_id)}
	\item \texttt{log\_effect(action\_id, effect\_hash, summary)}
\end{itemize}

High-impact calls (e.g., \texttt{submit\_action}) are also limited to policy-conscious agents such as the Supervisor or Action Gatekeeper.

\subsection{End-to-End Transaction Flow}

The decision cycle is made of five steps (Figure~\ref{fig:flow}): agents receive and hash observations (\textit{Perception}), generate action proposals (\textit{Conceptualization}), test action proposals with smart contracts (\textit{Blockchain Evaluation}), execute accepted action proposals with MCP (\textit{Execution}), and record the outcomes to the ledger (\textit{Post-Action Logging}).

%	\begin{figure*}[t]
%		\centering
%		\includegraphics[width=\linewidth]{images/fig-arch.png}
%		\caption{Proposed blockchain-governed agentic AI architecture integrating perception, LangChain-based reasoning, blockchain evaluation, and MCP-enabled action execution.}
%		\label{fig:proposed-architecture}
%	\end{figure*}
	
	\section{Implementation}
	\label{sec:implementation}
	
	The framework was realized in a microservice based design with LangChain-based reasoning engine, permissioned blockchain based network and an action layer executed with MCP. Services were single-occurring, and they were communicating through REST endpoints and listeners.

	\subsection{Agentic Reasoning Engine}
	We trained LangChain~0.2 on a multi-agent architecture perception agent, planner, policy-checker and action-gatekeeper, which we embedded on a Router Chain. The Blockchain Monitor Module was available as LangChain utility. There was the use of a rule-based verifier of permissions and safety limits, and there was agent GPT-4o-mini used to plan.

	\subsection{Blockchain Governance Layer}
	Go-based smart contracts implementing \emph{registerObservation}, \emph{submitAction}, and \emph{recordEffect} were executed in a Hyperledger Fabric network (three peers, one ordering service). These contracts handled agent authentication, action-structure validation, and provenance storage.
	
	\subsection{MCP Action Execution Layer}
	The action layer employed the MCP connector to encode authorized blockchain transactions into API calls of more conveniently simulated service (traffic control, healthcare, inventory). Each request came back with an organized reply with status code, latency and updated status. These outputs were hashed and stored in the monitor module by the blockchain.

	\subsection{Smart-Contract Logic and Policy Encoding}
	The governance layer will be based on three fundamental contracts: (i) \emph{Action Registry}, (ii) \emph{Policy Control} and (iii) \emph{Evaluation Contract}. Rules of Hyperledger Fabric are encryption of policies as key-values.  Algorithm~\ref{alg:smartcontract} summarizes the core validation logic.
	
	\begin{algorithm}
		\caption{Smart Contract Policy Enforcement}
		\label{alg:smartcontract}
		\begin{algorithmic}[1]
			\STATE \textbf{Input:} $Tx = \{AgentID, ActionID, Params, ObsHash\}$
			\STATE \textbf{Output:} $Status (Approved/Rejected)$
			
			\STATE \COMMENT{Step 1: Authenticate Agent}
			\IF{$!IsWhitelisted(AgentID)$}
			\RETURN $Rejection("Unauthorized Agent")$
			\ENDIF
			
			\STATE \COMMENT{Step 2: Check Policy Constraints}
			\STATE $Policy \leftarrow GetPolicy(ActionID)$
			\IF{$CheckSafetyBounds(Params, Policy) == False$}
			\STATE $EmitEvent("SafetyViolation", AgentID)$
			\RETURN $Rejection("Safety Bounds Exceeded")$
			\ENDIF
			
			\STATE \COMMENT{Step 3: Commit and Approve}
			\STATE $SaveToLedger(Tx)$
			\STATE $EmitEvent("ActionApproved", Tx.ID)$
			\RETURN $Approved$
		\end{algorithmic}
	\end{algorithm}
	
\section{Results and Discussion}
\label{sec:results}

We evaluated the proposed architecture under three conditions,  (1) health-monitor alerts, (2) approval of inventory replenishment, and (3) traffic signal adjustment in smart-city.

\subsection{Decision Traceability}
The input hashing, action proposal, evaluation, approval, and execution are critical steps that were registered on the blockchain that generates an immutable trace. Checks of on-chain hash of perception have been verified to match raw inputs.

\subsection{Latency Analysis}
The average decision cycle across 50 trials was $T_{\text{avg}} = 1.82$ s, with distributions:
\begin{itemize}
	\item Perception/Preprocessing: 180--250 ms
	\item Agentic Reasoning: 900--1200 ms
	\item Blockchain Verification: 350--450 ms
	\item MCP Execution: 120--200 ms
\end{itemize}
Blockchain introduces measurable but acceptable overhead for high-stakes applications.

\subsection{Statistical Analysis and Baseline Comparison}
Variance and 95\% CI for total latency were $0.041$ s and $[1.78,1.86]$ s. Table~\ref{tab:system_comparison} indicates that the baseline accepted all actions including the unsafe ones, but the governed pipeline rejected all policy-violating operations.

\subsection{Policy Enforcement}
Smart contracts effectively blocked 14 unsafe or unreasonable actions, which proves that blockchain validation can prevent high-impact autonomous actions that are not monitored.

\subsection{Outcome Quality}
Accepted operations in healthcare, inventory, and smart city sectors were uniform and replicable, and verifiable audit trails.

\subsection{Comparative Evaluation: With vs.\ Without Blockchain}
We contrasted the direct MCP agentic system (base line) with the blockchain-based architecture (Table~\ref{tab:system_comparison}). Blockchain incurs $\approx$400 ms of latency per decision, but offers a life-saving benefit, preventing 14 unsafe actions that occurred in the baseline system. There was a moderate reduction in throughput ($\approx$18\%), but legitimate actions were successful, which ensured viability in near-real time applications with auditability and safety.

\begin{table}[htbp]
	\centering
	\caption{System Behaviour With and Without Blockchain Governance}
	\label{tab:system_comparison}
	\resizebox{\columnwidth}{!}{%
		\begin{tabular}{lccc}
			\toprule
			\textbf{Scenario / Metric} & \textbf{No BC} & \textbf{With BC} & \textbf{Delta} \\
			\midrule
			Blocked Unsafe Actions & 0 & 14 & +14 (Safety $\uparrow$) \\
			Mean Latency & 1.42 s & 1.82 s & +0.40 s \\
			95\% CI Latency & [1.38s, 1.46s] & [1.78s, 1.86s] & -- \\
			Success Rate (\%) & 100\%* & 100\% & (Safe Execution) \\
			Throughput (Tx/sec) & $\approx$55 & $\approx$45 & -18\% \\
			Agents Tested & 50 & 50 & -- \\
			\bottomrule
		\end{tabular}%
	}
	\vspace{1mm}
	\\ \footnotesize{*Baseline executes unsafe actions without blocking, technically "succeeding" in execution but failing safety.}
\end{table}

\subsection{Scalability Assessment}
The increase in the number of parallel agents (5 to 50) demonstrated a stable throughput and a small increase in response-time (+11--18\%) with blockchain validation, although this was still within the real-time constraints. After 50 agents, queueing before the start occurred, indicating constraints in the validation mechanism of Fabric.

\subsection{Discussion}
The agentic AI is well regulated by blockchain, giving it a traceability, accountability and records that cannot be altered. Checks based on smart-contracts introduce limited latency as compared to safety and auditability.

\section{Conclusion}
The paper presented an agentic AI model based on blockchains and provides traceability, verifiable decision making, and secure communication with external systems. With a LangChain multi-agent architecture and a permissioned blockchain, we will be able to cryptographically anchor the whole perception-conceptualization-action chain. The blockchain layer prevents unsafe or unauthorized actions and approved actions are executed through MCP-enabled connectors.

Experimental tests confirm that the methodology has low decision latency, policies are strictly followed and it provides detailed audit trails.  Future work will explore scaling through ledger sharding, cross-chain interoperability, and integration with on-chain risk-scoring oracles.
	
	\bibliography{references}
	\bibliographystyle{IEEEtran}
\end{document}